\documentclass[10pt,twocolumn,letterpaper]{article}

\usepackage{iccv}
\usepackage{times}
\usepackage{epsfig}
\usepackage{graphicx}
\usepackage{amsmath}
\usepackage{amssymb}
\usepackage{etoolbox}
\usepackage{array}
\usepackage{setspace}

\usepackage[pagebackref=true,breaklinks=true,letterpaper=true,colorlinks,bookmarks=false]{hyperref}

\iccvfinalcopy 

\def\iccvPaperID{2344} 

\ificcvfinal\fi
\begin{document}

\title{Object-Aware Instance Labeling for Weakly Supervised Object Detection}

\author{
\and
Satoshi Kosugi ~~~~Toshihiko Yamasaki ~~~~Kiyoharu Aizawa\\
The University of Tokyo, Japan\\
{\tt\small \{kosugi, yamasaki, aizawa\}@hal.t.u-tokyo.ac.jp}
\and
}

\maketitle
\ificcvfinal\thispagestyle{empty}\fi

\begin{abstract}
  \vspace{-1mm}
  Weakly supervised object detection (WSOD), where a detector is trained with only image-level annotations, is attracting more and more attention.
  As a method to obtain a well-performing detector, the detector and the instance labels are updated iteratively.
  In this study, for more efficient iterative updating, we focus on the instance labeling problem, a problem of which label should be annotated to each region based on the last localization result.
  Instead of simply labeling the top-scoring region and its highly overlapping regions as positive and others as negative, we propose more effective instance labeling methods as follows.
  First, to solve the problem that regions covering only some parts of the object tend to be labeled as positive, we find regions covering the whole object focusing on the context classification loss.
  Second, considering the situation where the other objects contained in the image can be labeled as negative, we impose a spatial restriction on regions labeled as negative.
  Using these instance labeling methods, we train the detector on the PASCAL VOC 2007 and 2012 and obtain significantly improved results compared with other state-of-the-art approaches.
\vspace{-4mm}
\end{abstract}

\begin{figure*}[t]
 \centering
 {\includegraphics[width=1\hsize]{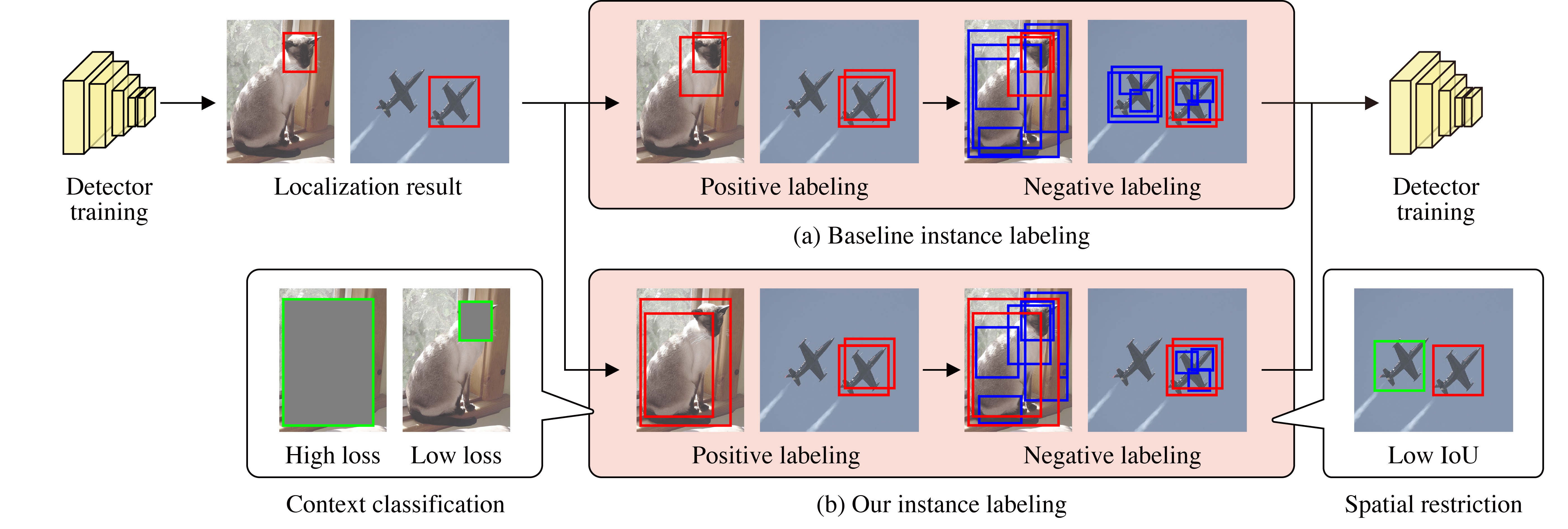}}
 \caption{Comparison of (a) baseline instance labeling and (b) our instance labeling.
 In the localization result, only the top-scoring region is shown (red box).
 Red/blue boxes in the instance labeling denote regions labeled as positive/negative.
 Our labeling method can label regions covering the whole object as positive focusing on the context classification loss
 and avoid labeling other objects as negative introducing a spatial restriction.\label{fig:instancelabeling}
 }
 \vspace{-3mm}
\end{figure*}

\begin{spacing}{0.99}
\section{Introduction}
Object detectors trained on large-scale datasets with instance-level annotations
({\it i.e.,} strongly supervised object detectors) have made significant progress~\cite{girshick2015fast,liu2016ssd,redmon2016you}
with the recent development of convolutional neural networks (CNNs),
but such detailed large-scale datasets are time-consuming and labor-intensive to collect accurately.
On the other hand, image-level labels that indicate the presence of an object can be acquired easily and in large amounts
because such labels take less time to annotate manually or can be collected using an image search on the Internet.
In order to take advantage of readily available image-level annotations, in this study,
we focus on the problem of training a detector with only image-level annotations; that is, weakly supervised object detection (WSOD).

As a method to obtain a well-performing detector with only image-level annotations,
both a detector and instance labels are updated iteratively.
Conventional methods include an alternating iterative strategy~\cite{cinbis2017weakly,jie2017deep,li2016weakly,song2014learning}.
A detector is trained on instance labels initialized based on a simple rule ({\it e.g.}, supposing the object is at the center of the image~\cite{song2014learning}),
and the instance labels are updated using the trained detector, over and over again.
Although the initial instance labels are rough and the detector trained on the initial labels has low performance,
the detector and the instance labels are refined step by step through the alternate optimization.
In a recent method~\cite{tang2017multiple}, for end-to-end iterative update of detectors and instance labels, multiple instance classifiers (object detectors) have been employed.
Each instance classifier is trained using the last instance classifier's localization result as supervisions.
The method can reduce the training time by optimizing multiple instance classifiers end-to-end, and achieves good performance.

In order to obtain efficient iterative update, we focus on an instance labeling problem, a problem of which label should be annotated to each region based on the last localization result.
The instance labeling method employed by one of the state-of-the-art methods~\cite{tang2017multiple} is rather simple;
the most confident region and its highly overlapping regions are labeled as positive, and other regions are labeled as negative or background.
For more effective instance labeling,~we propose two methods: Context-Aware Positive~(CAP) labeling and Spatially Restricted Negative (SRN) labeling.
CAP labeling is aimed at solving the problem that the most discriminative parts of the object ({\it e.g.}, faces in the person class) tend to be detected rather than the whole object.
We find that the classification loss of the context of the region ({\it i.e.}, the outside of the region) differs depending on whether the region covers the whole object or not.
Utilizing this characteristic, we replace the incomplete detected region with a region covering the whole object.
In addition to CAP labeling, we develop SRN labeling to consider the negative labeling; that is, which region should be annotated as background.
When an image has multiple objects of the same class, even though one object is labeled as positive, the other objects can be labeled as negative.
SRN labeling solves this problem by imposing a spatial restriction on negative labeling.
We show a comparison with the baseline labeling in Figure~\ref{fig:instancelabeling}.
\end{spacing}

In order to verify the efficacy of our method, we conducted experiments on the PASCAL VOC 2007 and 2012 datasets~\cite{everingham2015pascal}.
The obtained mean Average Precision (mAP) scores are 47.6\% and 43.4\% respectively, which surpasses other state-of-the-art methods.

In summary, the contributions of this paper are as follows:
\begin{itemize}
  \setlength{\parskip}{0cm}
  \setlength{\itemsep}{0cm}
  \item We improve the WSOD method from the viewpoint of instance labeling.
  \item We propose two methods for instance labeling. The first method is aimed at finding a region covering the whole object based on the context classification loss.
  The second can avoid labeling objects as negative by imposing a spatial restriction.
  \item Experiments on the PASCAL VOC 2007 and 2012 datasets demonstrate that our method can achieve better performance than other state-of-the-art approaches.
\end{itemize}

\section{Related works}
WSOD is a task where a detector is trained with only image-level annotations.
Methods for WSOD can be roughly divided into three approaches: the alternating approach, end-to-end approach, and transferring approach.

\subsection{Alternating approach}
A conventional method to train a detector with only image-level annotations is an alternating approach~\cite{cinbis2017weakly,jie2017deep,li2016weakly,song2014learning}.
Song et al.~\cite{song2014learning} initialized instance labels supposing the object is at the center of the image and trained the detector.
The initial instance labels are rough, because the location information of the object is unavailable, and the detector trained on the initial labels has low performance.
By updating the detector and the instance labels alternately, the detector and the instance labels are refined step by step.

Based on the alternating approach, other methods were developed to detect objects more accurately.
Li et al.~\cite{li2016weakly} trained a classifier using entire images,
and then selected confident class-specific region proposals using a mask-out strategy.
Cinbis et al.~\cite{cinbis2017weakly} developed a multi-fold learning method
to solve the problem that alternating approaches are easily trapped in local optima.
Jie et al.~\cite{jie2017deep} developed a self-taught learning method to select more reliable seed positive proposals.
Our instance labeling method can be applied to these alternating approaches,
but alternating approaches, which split the training process between the optimizing detector and updating instance labels,
tend to get stuck in local optima and are time-consuming.
Therefore, we apply our instance labeling method to an end-to-end iterative approach described below.

\vspace{-1mm}
\subsection{End-to-end approach}
\vspace{-1mm}
Bilen et al.~\cite{bilen2016weakly} proposed a weakly supervised deep detection network (WSDDN) with two streams: a classification stream and a detection stream.
The outputs of these two streams are combined and used to score each region.
Kantorov et al.~\cite{kantorov2016contextlocnet} extended WSDDN to consider contextual information.
Diba et al.~\cite{diba2017weakly} and Wei et al.~\cite{wei2018ts2c} used semantic segmentation based on class activation map~\cite{zhou2016learning} to discover region proposals that tightly cover the object.
Tang et al.~\cite{tang2018weakly} developed high-quality region proposals by exploiting the low-level information in CNN.

An end-to-end approach (online instance classifier refinement, OICR) that takes advantage of alternating approaches was proposed by Tang et al.~\cite{tang2017multiple}.
OICR takes WSDDN as the initial instance localization method and has multiple instance classifiers (object detectors).
The first instance classifier is trained on the instance-level supervisions labeled by WSDDN,
and the second instance classifier is trained using the localization result of the first instance classifier as supervisions.
Similar to alternating approaches, the instance classifiers and the instance labels are refined iteratively.
Because OICR takes less time to train and has higher performance than alternating approaches,
recent methods~\cite{tang2018weakly,wei2018ts2c} employ OICR as a baseline.
We also apply our instance labeling method to OICR.

\subsection{Transferring approach}
The location information obtained by the above WSOD method can be transferred to a supervised object detector.
Shen et al.~\cite{shen2018generative} proposed a generative adversarial learning paradigm.
They introduced a discriminator and trained a one-stage detector similar to SSD~\cite{liu2016ssd} so that the discriminator cannot distinguish the detector and OICR~\cite{tang2017multiple} model.
The trained one-stage detector achieves faster detection.
Zhang et al.~\cite{zhang2018w2f} proposed pseudo labeling methods named pseudo ground-truth excavation and pseudo ground-truth adaptation.
Using these methods, they generated pseudo ground truth boxes from localization result of OICR~\cite{tang2017multiple} and trained a faster R-CNN~\cite{ren2015faster} model.
Zhang et al.~\cite{zhang2018zigzag} proposed a zigzag learning strategy, in which
they developed a criterion (the Energy Accumulation Score) to automatically measure and rank localization difficulty.
As the localization result of WSOD is unreliable, at first they used easy images to localize and added difficult images progressively.
Instead of only using the top-scoring regions as pseudo ground truth,
supervised object detectors can be trained more effectively with these transferring approaches.
We can obtain a further performance improvement combining transferring approaches and our localization result.


\section{Method}
A goal of WSOD is to train a detector with only image-level annotations.
As a typical method to obtain a well-performing detector, both the detector and instance labels are updated iteratively.
In order to train a detector iteratively, we have to solve the problem of which label should be annotated to each region based on the last localization result.
In this study, we focus on this problem; that is, the instance labeling problem.

Among iteratively updating methods, we employ the OICR~\cite{tang2017multiple} as a baseline.
We first introduce OICR shortly.

\noindent
\newline
{\bf OICR~~}
OICR includes two modules, multiple instance classification and instance refinement.
In particular, an end-to-end WSOD method named WSDDN~\cite{bilen2016weakly} is employed as a multiple instance classification module.
WSDDN includes two streams that calculate region-wise scores in a different way based on CNN features pooled by Spatial Pyramid Pooling (SPP)~\cite{he2015spatial},
a classification stream, and a detection stream.
The classification stream conducts a softmax operation on each region proposal for classification.
The detection stream conducts a softmax operation on each class in order to estimate which region is most valuable for classification.
Both output scores are combined by an element-wise product and defined as each region's detection score.

Suppose an input image is $X$, the image label vector is ${\mathbf{Y}} = [y_1, ..., y_C]$,
and its region proposals by Selective Search~\cite{uijlings2013selective} are $\{r_1, r_2, ..., r_J\}$,
where $C$ denotes the number of image classes, $y_c=1$ or $0$ denotes the image with or without object $c$, and $J$ denotes the number of the region proposals.
Through WSDDN, we obtain the initial proposal score matrix ${\mathbf{x}}^0 \in \mathbb{R}^{C\times J}$,
where each element $x^0_{cj}$ denotes region $r_j$'s score for class $c$.
When WSDDN is trained, the image score $\phi_c$ is obtained by the sum over all proposals, $\phi_c$ = $\sum_{j=1}^{J} x^0_{cj}$, and the following multi class cross entropy is minimized,
\begin{equation}
{L}_{b}  = - \sum_{c=1}^{C}\{y_{c}\log \phi_{c} + (1-y_{c})\log (1-\phi_{c})\}.\label{eq:wsddn_loss}
\end{equation}

By utilizing WSDDN as an initial localization network,
multiple instance classifiers are trained progressively to refine the localization result and obtain a well-performing detector.
Here, let $K$ be a number of instance classifiers and ${\mathbf{x}}^k \in \mathbb{R}^{(C+1)\times J}$ be an output proposal score of the $k^{\rm th}$ instance classifier.
Different from ${\mathbf{x}}^0$, ${\mathbf{x}}^k (k \in \{1,...,K\})$ has the $\{C+1\}^{\rm th}$ dimension for background.
To train the multiple instance classifiers progressively,
the ground truth label ${\mathbf{y}}^k \in \mathbb{R}^{(C+1)\times J}$ for the $k^{\rm th}$ instance classifier is made from the last instance classifier's output ${\mathbf{x}}^{k-1}$.
Based on ${\mathbf{y}}^k$, each instance classifier is trained to minimize the following loss:
\begin{equation}
 {L}^{k}_{r}  = -\frac{ 1 }{J} \sum_{j=1}^{J}\sum_{c=1}^{C+1}y_{cj}^{k}\log x_{cj}^{k}.\label{eq:loss}
\end{equation}

In OICR, instance labeling is a problem of how to generate an instance label ${\mathbf{y}}^k$ from the last localization result ${\mathbf{x}}^{k-1}$.
Suppose an image $X$ has class label c, they first select proposal $r_{j_{c}}$ with the highest score,
\begin{equation}
j_c  =  {\rm arg} \max_j x_{cj}^{k-1},
\end{equation}
and inspired by the fact that highly overlapped regions should have the same label, they formulate the following labeling algorithm,
\begin{equation}
  \begin{split}
  y_{cj}^{k} = \begin{cases}
    1 & {\rm if~}{\rm IoU}(r_j, r_{j_{c}}) > I_t \\
    0 & {\rm otherwise}\label{eq:positivelabeling}
  \end{cases},
\end{split}
\end{equation}
where ${\rm IoU}$ is a function of calculating Intersection over Union (IoU) between two regions and $I_t$ is a threshold.
When multiple classes satisfy ${\rm IoU}(r_j, r_{j_c}) > I_t$, $y_{cj}^{k}$ whose $c = {\rm arg} \max_{c'} {\rm IoU}(r_j, r_{j_{c'}})$ is $1$ and the others are $0$.
If a region is not assigned any object classes, that is, $y_{cj}^{k}$ is $0$ for all $c\in \{1, ..., C\}$,
the region is labeled as background,
\begin{equation}
  y_{(C+1)j}^{k} = 1.
\end{equation}
However, the label generated from the last localization result is unreliable, especially at the beginning of the training.
This results in the instability of the training.
To solve this problem, the loss function in Eq.~(\ref{eq:loss}) is changed to the weighted version as follows,
\begin{equation}
w_{j}^{k} = x^{k-1}_{{c}j_{c}},\label{eq:weight}
\end{equation}
\begin{equation}
{L}_{r}^{k}  = -\frac{ 1 }{J} \sum_{j=1}^{J}\sum_{c=1}^{C+1}w_{j}^{k}y_{cj}^{k}\log x_{cj}^{k}.
\end{equation}
When an image has multiple classes, in Eq.~(\ref{eq:weight}), $c = {\rm arg} \max_{c'} {\rm IoU}(r_j, r_{j_{c'}})$.
At the beginning of the training or for a difficult image to localize,
the weight $w_{j}^{k}$ takes a low value and the contribution to the training becomes small.

\noindent
\newline
{\bf Problem~~}
The simple instance labeling method described above has two problems.
First, the most discriminative part of the object tends to be detected rather than the whole object.
If $r_{j_c}$ does not highly overlap the whole object,
the progressive update is trapped in the local optima.
Second, the simple instance labeling does not take into account cases where an image contains multiple objects of the same class.
Even though one object is correctly labeled as positive, other objects can be incorrectly labeled as background.
To solve these problems, we propose more effective instance labeling methods named CAP labeling and SRN labeling.


\begin{figure*}[t]
 \centering
 {\includegraphics[scale=0.28]{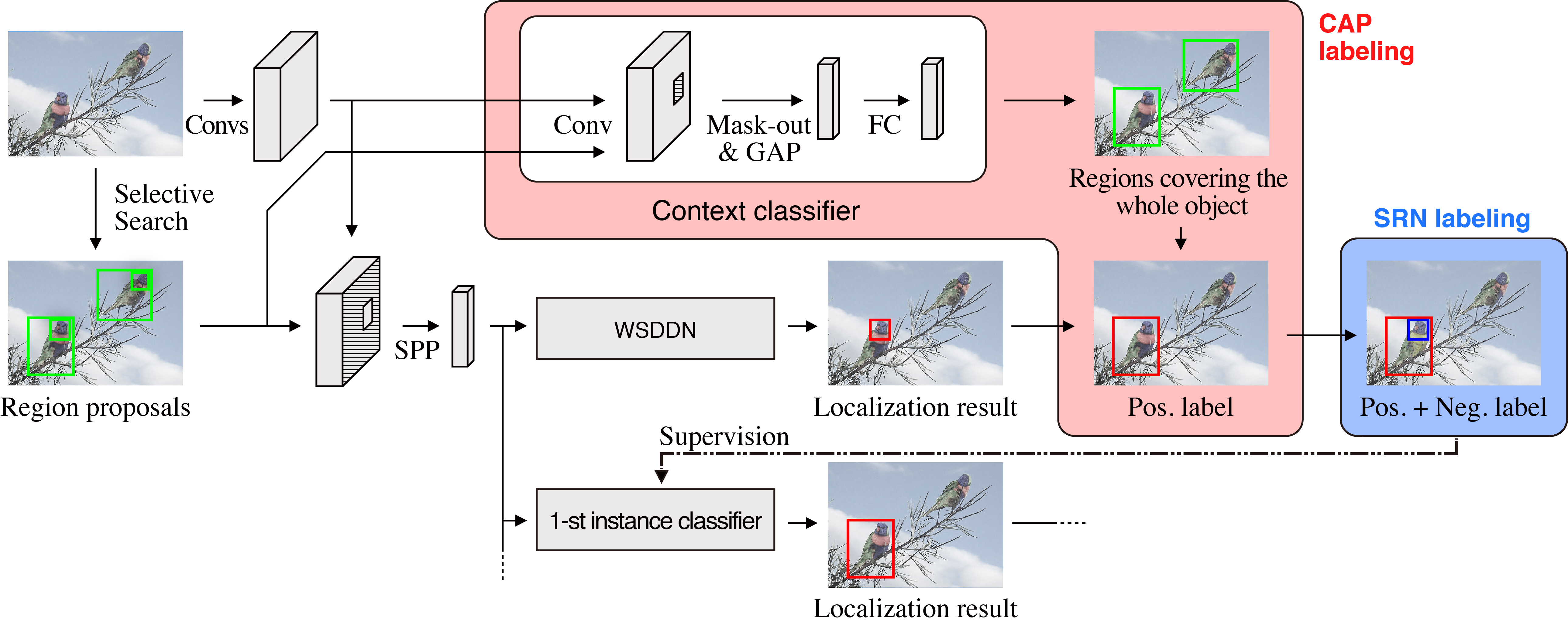}}
 \caption{Overview of our method.
 We label each region as positive or negative based on the last localization result,
 and the next instance classifier is trained on these instance labels.
 To label regions covering the whole object as positive, we discover such regions focusing on the context classification loss.
 In addition, we employ a spatial restriction to avoid labeling the other objects as negative.
 Although this image shows only the first instance classifier, the second and subsequent instance classifiers are trained in the same way.
 \label{fig:architecture}
 \vspace{-2mm}
 }
\end{figure*}

\subsection{CAP Labeling}
We propose CAP labeling to avoid the problem that the most discriminative part of the object tends to be detected rather than the whole object.
We introduce a network that judges whether a region covers the whole object or not,
and when we generate the instance labels, the top-scoring region is selected from regions covering the whole object.

In some previous methods~\cite{bazzani2016self,li2016weakly}, mask-out strategy is used to find the whole object.
If a mask-out image by a region drops the classification confidence, that region can be considered discriminative.
However, we experimentally find that the mask-out by regions covering only some parts of the object can drop the classification confidence.
Such a mask-out method is improper for discovery of regions covering the whole object.

In order to judge whether a region covers the whole object or not more properly, we focus on the research of Tanaka et al.~\cite{tanaka2018joint},
who deal with a problem of classification with noisy labels, where a classifier is trained with noisy labeled images.
Here, noisy labeled images mean incorrectly labeled images ({\it e.g.}, a dog image labeled as a cat).
According to them, when a classifier is trained on noisy labeled images,
the training loss differs depending on whether the data is noisy or clean. The loss tends to decrease for clean images and is hard to decrease for noisy labeled images.

We find that this characteristic can be used to judge whether a region covers the whole object or not.
We focus not on the inside of the region but on the outside of the region.
We call the outside of the region as the context of that region.
Taking an image containing a cat as an example;
when a region covers the whole cat, no cat exists in the context of the region.
On the other hand, when a region does not cover the whole cat, some parts of the cat are in the context.
If we label the contexts of all regions as a cat, these are noisy labeled images:
when a region covers the whole cat, the context of that region is noisy and otherwise clean.
By training a classifier using this data, classification loss differs depending on whether a region covers the whole cat or not.

As a simple method to train a classifier based on the context,
the inside of the region is filled with the mean pixel value before the image is input into a CNN.
However, this method requires CNN forwarding for each region and is time-consuming.
To achieve low computational cost, we perform mask-out to the CNN feature.
The CNN feature corresponding to the inside of the region is filled with zero values.
Then the feature after mask-out is pooled with global average pooling~(GAP) and input to a fully connected~(FC) layer.

Let the output of the classifier with CNN feature mask-out be ${\mathbf{p}} \in \mathbb{R}^{C\times J}$,
where each element $p_{cj}$ denotes a probability for class $c$ of $r_j$'s context.
The classifier is trained to minimize the standard multi class cross entropy loss with the image-level label ${\mathbf{Y}}$,
\begin{equation}
{L}_{context}  = - \frac{ 1 }{J}\sum_{j=1}^{J}\sum_{c=1}^{C}\{y_{c}\log p_{cj} + (1-y_{c})\log (1-p_{cj})\}.\label{eq:context_loss}
\end{equation}
If a region covers the whole object, the training loss of the context is high after training because the context of the region is noisy.
In other words, the class probability $p_{cj}$, whose class $c$ is contained by the image ($y_{c}=1$), is low.
On the other hand, if a region does not cover the whole object, the class probability of the context, which is clean, is high.

When we conduct instance labeling, the top-scoring region is selected from regions whose context class probabilities are low,
\begin{equation}
  j_c  =  {\rm arg} \max_j x_{cj}^{k-1}~~~~~~ s.t.~~  p_{cj} < P_t,\label{eq:topscoring2}
\end{equation}
where $P_t$ is a threshold. Then following the OICR method, highly overlapping regions are labeled as positive based on Eq.~(\ref{eq:positivelabeling}).

Even though a region is covering the whole object, the training loss of the context can decrease in some cases; for example,
when the context is closely related to the object ({\it e.g.}, an aeroplane and sky) or when two or more objects are in an image.
To solve this problem, we introduce Xiao et al.'s~\cite{xiao2018deep} saliency map.
Following the previous WSOD method of Wei et al.~\cite{wei2018ts2c},
we define areas whose saliency is higher than 0.06 as foreground and others as background.
When the classifier is trained based on Eq.~(\ref{eq:context_loss}), background areas are filled with the mean pixel value before it is input to the classifier.
When the class probability is calculated for Eq.~(\ref{eq:topscoring2}), we divide foreground segments to each independent segment,
select the foreground segment that has the highest IoU between the segment and the box, and fill the other areas with the mean pixel value.
As a result, the object on which the box is focusing is visible and the other objects are hidden.

\subsection{SRN Labeling}
In CAP labeling, we label regions highly overlapped by $r_{j_c}$ as positive.
If a region is not assigned any object class, that is, $y_{cj}^{k}$ is $0$ for all $c\in \{1, ..., C\}$,
the region is labeled as background, $y_{(C+1)j}^{k} = 1$.
This labeling has a problem:
when an image has multiple objects of a specific class, even though one object is correctly labeled as positive,
the other objects are labeled as background.

To solve this problem, we propose SRN labeling.
This method is inspired by the fact that at a distant area from the object other objects may exist.
In SRN labeling, we put a spatial restriction on regions that are trained as background by modifying the weight in Eq.~(\ref{eq:weight}) as follows,
\begin{equation}
  w_{j}^{k} = \begin{cases}
    x^{k-1}_{cj_c} & {\rm if~~} {\rm IoU}(r_j, r_{j_c}) > i_t \\
    0 & {\rm otherwise}
  \end{cases},
\end{equation}
where $i_t$ is a threshold that is lower than $I_t$.

Originally, $w_{j}^{k}$ is aimed at restricting the contribution of unreliable labels, such as those generated at the beginning of the training.
SRN labeling is the spatial version of this: we regard labels at remote areas as unreliable.

\subsection{Overall architecture}
The overall architecture is shown in Figure \ref{fig:architecture}.
When training, we first train the context classifier according to the loss in Eq.~(\ref{eq:context_loss}) and calculate the context class probability $p_{cj}$.
Then we train the WSDDN and the multiple instance classifiers to minimize the following loss,
\begin{equation}
{L}_{OICR}  = {L}_{b} + \sum_{k=1}^{K}{L}_{r}^{k}.
\end{equation}
For the test, we ignore the context classifier and the WSDDN and take an average of multiple instance classifiers' output to obtain the final detection results.

\newcolumntype{C}{>{\centering\arraybackslash}p{5mm}}
\begin{table*}[t]
  \centering
  \caption{Average precision (\%) on PASCAL VOC 2007 and 2012 test datasets.\vspace{-3mm}}
    {\tabcolsep=1.12mm
    {\scriptsize
  \begin{tabular}{l|CCCCCCCCCCCCCCCCCCCC|C}
    \multicolumn{22}{l}{} \vspace{-1.5mm}\\\hline
        method\rule[0mm]{0mm}{3mm}&aero&bike&bird&boat&bottle&bus&car&cat&chair&cow&table&dog&horse&mbike&person&plant&sheep&sofa&train&tv&mAP\vspace{1mm}\\ \hline\hline
        - VOC 2007&\multicolumn{20}{l|}{}& \rule[0mm]{0mm}{3mm}\vspace{1mm}\\
    OICR~\cite{tang2017multiple} \rule[0mm]{0mm}{1mm}& 58.0 & 62.4 & 31.1 & 19.4 & 13.0 & 65.1 & 62.2 & 28.4 & 24.8 & 44.7 & 30.6 & 25.3 & 37.8 & 65.5 & 15.7 & 24.1 & 41.7 & 46.9 & 64.3 & {\bf 62.6} & 41.2\vspace{0mm}\\
    SGWSOD~\cite{lai2017saliency} \rule[0mm]{0mm}{3mm} & 48.4&61.5&33.3&{\bf 30.0}&15.3&{\bf 72.4}&62.4&59.1&10.9&42.3&34.3&{\bf 53.1}&48.4&65.0&20.5&16.6&40.6&46.5&54.6&55.1&43.5\vspace{0mm}\\
        ${\rm TS^2C}$~\cite{wei2018ts2c} \rule[0mm]{0mm}{3mm} & 59.3 & 57.5 &{\bf 43.7}& 27.3 & 13.5 & 63.9 & 61.7 & 59.9 & 24.1 & 46.9 & 36.7 & 45.6 & 39.9 & 62.6 & 10.3 & 23.6 & 41.7 & 52.4 & 58.7 & 56.6 & 44.3\vspace{1mm}\\
        WSRPN~\cite{tang2018weakly} & 57.9&{\bf 70.5}&37.8&5.7&{\bf 21.0}&66.1&{\bf 69.2}&59.4&3.4&{\bf 57.1}&{\bf 57.3}&35.2&{\bf 64.2}&{\bf 68.6}&{\bf 32.8}&{\bf 28.6}&{\bf 50.8}&49.5&41.1&30.0&45.3\vspace{1mm}\\\hline
        Ours \rule[0mm]{0mm}{3mm} &{\bf 61.5}&64.8&{\bf 43.7}&26.4&17.1&67.4&62.4&{\bf 67.8}&{\bf 25.4}&51.0&33.7&47.6&51.2&65.2&19.3&24.4&44.6&{\bf 54.1}&{\bf 65.6}&59.5&{\bf 47.6}\vspace{1mm}\\\hline\hline
        - VOC 2012&\multicolumn{20}{l|}{}& \rule[0mm]{0mm}{3mm}\vspace{1mm}\\
    OICR~\cite{tang2017multiple} \rule[0mm]{0mm}{1mm}&-&-&-&-&-&-&-&-&-&-&-&-&-&-&-&-&-&-&-&-&37.9\vspace{0mm}\\
    SGWSOD~\cite{lai2017saliency} \rule[0mm]{0mm}{3mm} & 51.7&61.0&32.3&20.4&{\bf 24.8}&{\bf 59.9}&45.2&62.2&13.7&45.1&13.6&51.0&51.2&64.9&{\bf 22.1}&21.2&39.9&19.1&44.3&49.1&39.6\vspace{0mm}\\
        ${\rm TS^2C}$~\cite{wei2018ts2c} \rule[0mm]{0mm}{3mm} &67.4&57.0&37.7&23.7&15.2&56.9&49.1&{\bf 64.8}&15.1&39.4&{\bf 19.3}&48.4&44.5&67.2&2.1&{\bf 23.3}&35.1&{\bf 40.2}&{\bf 46.6}&45.8&40.0\vspace{1mm}\\
        WSRPN~\cite{tang2018weakly} &-&-&-&-&-&-&-&-&-&-&-&-&-&-&-&-&-&-&-&-&40.8\vspace{1mm}\\\hline
        Ours \rule[0mm]{0mm}{3mm} &{\bf 70.2}&{\bf 61.3}&{\bf 43.8}&{\bf 28.9}&23.5&54.0&{\bf 52.1}&55.2&{\bf 19.1}&{\bf 51.0}&15.6&{\bf 52.6}&{\bf 56.6}&{\bf 68.9}&22.0&21.7&{\bf 43.6}&37.0&34.8&{\bf 56.3}&{\bf 43.4}\vspace{1mm}\\\hline
        \multicolumn{22}{l}{} \vspace{0mm}\\
  \end{tabular}
  }
    \label{map}
    }
    \vspace{-3mm}
\end{table*}

\begin{table*}[t]
  \centering
  \caption{CorLoc (\%) on PASCAL VOC 2007 and 2012 trainval datasets.}
    {\tabcolsep=1.12mm
    {\scriptsize
  \begin{tabular}{l|CCCCCCCCCCCCCCCCCCCC|C}
    \multicolumn{22}{l}{} \vspace{-1.5mm}\\\hline
        method\rule[0mm]{0mm}{3mm}&aero&bike&bird&boat&bottle&bus&car&cat&chair&cow&table&dog&horse&mbike&person&plant&sheep&sofa&train&tv&mean\vspace{1mm}\\ \hline\hline
        - VOC 2007&\multicolumn{20}{l|}{}& \rule[0mm]{0mm}{3mm}\vspace{1mm}\\
    OICR~\cite{tang2017multiple} \rule[0mm]{0mm}{1mm}& 81.7&80.4&48.7&49.5&32.8&81.7&85.4&40.1&40.6&79.5&35.7&33.7&60.5&88.8&21.8&57.9&76.3&59.9&75.3&{\bf 81.4}&60.6\vspace{0mm}\\
        ${\rm TS^2C}$~\cite{wei2018ts2c} \rule[0mm]{0mm}{3mm} & 84.2&74.1&61.3&52.1&32.1&76.7&82.9&66.6&42.3&70.6&39.5&57.0&61.2&88.4&9.3&54.6&72.2&60.0&65.0&70.3&61.0\vspace{0mm}\\
    SGWSOD~\cite{lai2017saliency} \rule[0mm]{0mm}{3mm} & 71.0&76.5&54.9&49.7&{\bf 54.1}&78.0&{\bf 87.4}&68.8&32.4&75.2&29.5&58.0&67.3&84.5&41.5&49.0&{\bf 78.1}&60.3&62.8&78.9&62.9\vspace{0mm}\\
        WSRPN~\cite{tang2018weakly} \rule[0mm]{0mm}{3mm} & 77.5&{\bf 81.2}&55.3&19.7&44.3&80.2&86.6&69.5&10.1&{\bf 87.7}&{\bf 68.4}&52.1&{\bf 84.4}&{\bf 91.6}&{\bf 57.4}&{\bf 63.4}&77.3&58.1&57.0&53.8&63.8\vspace{0mm}\\
    Teh et al.~\cite{teh2016attention} \rule[0mm]{0mm}{3mm} &84.0&64.6&{\bf 70.0}&{\bf 62.4}&25.8&80.7&73.9&71.5&35.7&81.6&46.5&{\bf 71.3}&79.1&78.8&56.7&34.3&69.8&56.7&77.0&72.7&64.6\vspace{1mm}\\\hline
        Ours \rule[0mm]{0mm}{3mm} &{\bf 85.5}&79.6&68.1&55.1&33.6&{\bf 83.5}&83.1&{\bf 78.5}&{\bf 42.7}&79.8&37.8&61.5&74.4&88.6&32.6&55.7&77.9&{\bf 63.7}&{\bf 78.4}&74.1&{\bf 66.7}\vspace{1mm}\\\hline\hline
        - VOC 2012&\multicolumn{20}{l|}{}& \rule[0mm]{0mm}{3mm}\vspace{1mm}\\
    OICR~\cite{tang2017multiple} \rule[0mm]{0mm}{1mm}&-&-&-&-&-&-&-&-&-&-&-&-&-&-&-&-&-&-&-&-&62.1\vspace{0mm}\\
    SGWSOD~\cite{lai2017saliency} \rule[0mm]{0mm}{3mm} & 70.4&79.3&54.1&44.9&{\bf 56.8}&{\bf 89.8}&72.3&69.2&41.0&67.3&32.3&61.1&72.0&85.0&43.9&{\bf 56.4}&77.8&42.6&64.0&77.6&62.9\vspace{0mm}\\
        ${\rm TS^2C}$~\cite{wei2018ts2c} \rule[0mm]{0mm}{3mm} &79.1&{\bf 83.9}&64.6&50.6&37.8&87.4&74.0&{\bf 74.1}&40.4&80.6&{\bf 42.6}&53.6&66.5&{\bf 88.8}&18.8&54.9&{\bf 80.4}&{\bf 60.4}&{\bf 70.7}&79.3&64.4\vspace{1mm}\\
        WSRPN~\cite{tang2018weakly} &-&-&-&-&-&-&-&-&-&-&-&-&-&-&-&-&-&-&-&-&64.9\vspace{1mm}\\\hline
        Ours \rule[0mm]{0mm}{3mm} &{\bf 86.5}&82.1&{\bf 67.2}&{\bf 58.7}&48.9&80.5&{\bf 75.6}&62.3&{\bf 46.0}&{\bf 81.9}&40.0&{\bf 64.2}&{\bf 82.4}&88.2&{\bf 44.2}&53.5&78.1&54.7&56.7&{\bf 82.9}&{\bf 66.7}\vspace{1mm}\\\hline
  \end{tabular}
  \vspace{-1mm}
  }
    \label{corloc}
    }
\end{table*}

\begin{figure}[t]
 \centering
 {\includegraphics[width=\hsize]{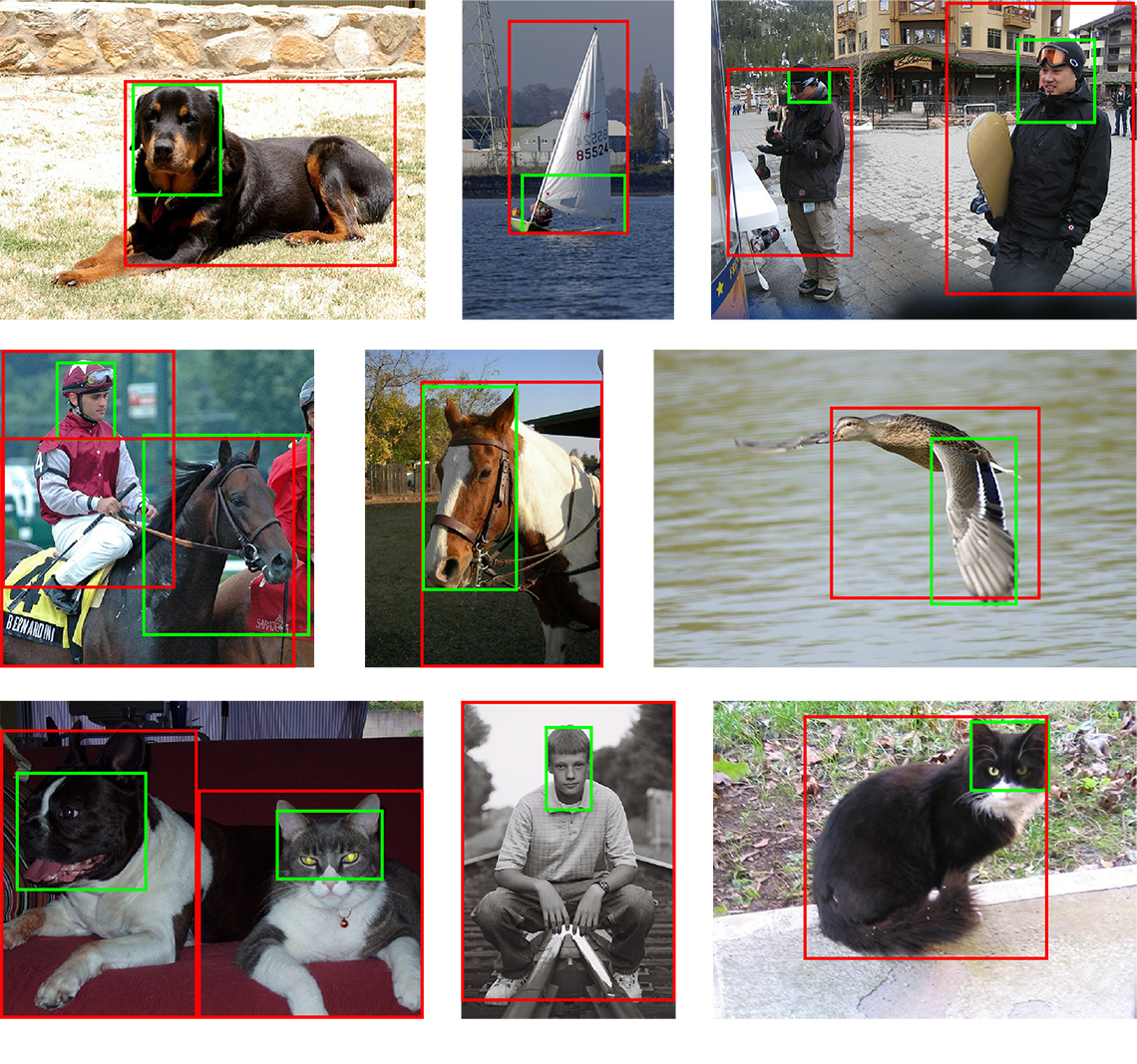}}
 \caption{Examples of detection results using our method and the baseline (OICR~\cite{tang2017multiple}).
 Red boxes denote the detection result of our method; green boxes denote the detection result of the baseline.\label{fig:detection_result}
 }
 \vspace{-3mm}
\end{figure}

\vspace{-1mm}
\section{Experiments}
We conducted experiments to verify the performance of our proposed method.

\subsection{Datasets and metrics}
To evaluate our method, we use the PASCAL VOC 2007 and 2012 datasets~\cite{everingham2015pascal}, which are standard benchmarks for WSOD.
These datasets have 9,962 and 22,531 images, respectively, with 20 classes and are divided into train, val, and test sets.
We select trainval images (5,011 for 2007 and 11,540 for 2012) to train our model with image-level annotations.
We employ two metrics to evaluate our method: mean Average Precision (mAP) and Correct Localization~(CorLoc). The
mAP is used to test the detection performance of our model on the test dataset, and CorLoc measures the localization accuracy on the trainval dataset.
Both metrics are based on the same IoU threshold between the predicted bounding boxes and ground truths, {\it i.e.}, ${\rm IoU} > 0.5$.

\vspace{2mm}
\subsection{Implementation}
Our model is built on the VGG16~\cite{simonyan2014very} model.
For the context classifier, a CNN feature is obtained replacing the last pooling layer and the fully connected layers with an additional convolutional
layer of size 3 $\times$ 3, stride 1, pad 1 with 1024 units following Zhou et al.~\cite{zhou2016learning}.
The feature corresponding inside of the bounding box is mask-out by zero, and the feature after mask-out is pooled by GAP followed by a FC layer.
For WSDDN~\cite{bilen2016weakly} and multiple instance classifiers~\cite{tang2017multiple}, the CNN features of VGG16 are pooled by the SPP layer extracting the feature corresponding to the inside of the bounding box.
During training, we first train the context classifier for 10K iterations (VOC 2007) or 20K iterations (VOC 2012) with the learning rate 0.001.
Then we train the WSDDN and the multiple instance classifiers for 70K iterations.
The learning rate is linearly increased from 0 to 0.001 for the first 10K iterations and fixed to 0.001 and 0.0001 for the following 30K iterations and the last 30K iterations, respectively.
The weight of the model is initialized with the one pretrained on the ImageNet~\cite{deng2009imagenet} dataset at the beginning of each training step.
Newly added layers are initialized using Gaussian distributions with means of 0 and standard deviations of 0.01.
Biases are initialized to 0.
The momentum is set to 0.9 and the weight decay is set to 0.0005.

As a region proposal method, we employ Selective Search~\cite{uijlings2013selective}, which generates about 2,000 proposals for each image.
For data augmentation, we use five scales \{480, 576, 688, 864, 1200\} resizing the shorter side to one of these scales,
and cap the longer side to 2000 with horizontal flips for both training and testing.
We set the number of instance classifiers $K$ to 3, and the mean output of these instance classifiers is used during testing.
Other parameters $I_t$, $P_t$, and $i_t$ are set to 0.5, 0.5, and 0.1 respectively.

\subsection{Comparisons with the state-of-the-arts}
\vspace{-2mm}
We compared our proposed method with previous methods based on a single VGG16 model.
The mAP result on VOC 2007 is shown in Table~\ref{map}.
This result shows that our method outperforms the other methods.
In particular, our method outperforms OICR~\cite{tang2017multiple} by 6.4\%, which is the baseline of our method.
This improvement is obtained by discovering regions covering the whole object and being aware of multiple objects.
Although OICR has a problem that only discriminative parts of cat and dog tend to be detected,
our method solves this problem, as shown by the gains of cat and dog (39.4\% and 22.3\% respectively).
In addition, ${\rm TS^2C}$~\cite{wei2018ts2c} and WSRPN~\cite{tang2018weakly} also employ OICR as the baseline, but our method outperforms these methods.
Examples of the detection results on the test dataset are shown in Figure~\ref{fig:detection_result}.
This result shows our method can effectively reduce false positive compared with OICR.
The mAP result on VOC 2012 is also shown in Table~\ref{map}.
The score of our method is higher than that of WSRPN~\cite{tang2018weakly}, which is another state-of-the-art method.
Table~\ref{corloc} shows the CorLoc result on VOC 2007 and 2012.
Our method outperforms each previous state-of-the-art method.

\newcolumntype{D}{>{\centering\arraybackslash}p{15mm}}
\begin{table}[t]
  \centering
  \caption{mAP (\%) on PASCAL VOC 2007 and 2012 test dataset by training FRCNN detectors.}
    {\tabcolsep=1.12mm
    {\scriptsize
  \begin{tabular}{l|DD}
    \multicolumn{3}{l}{} \vspace{0.1mm}\\\hline
        method\rule[0mm]{0mm}{3mm}&VOC~2007&VOC~2012\vspace{1mm}\\ \hline\hline
    OICR-Ens + FRCNN~\cite{tang2017multiple} \rule[0mm]{0mm}{3mm}& 47.0 & 42.5\vspace{0mm}\\
    ${\rm TS^2C}$ + FRCNN~\cite{wei2018ts2c} \rule[0mm]{0mm}{3mm}& 48.0 & 44.4\vspace{0mm}\\
    WSRPN-Ens + FRCNN~\cite{tang2018weakly} \rule[0mm]{0mm}{3mm}& 50.4 & 45.7\vspace{1mm}\\\hline
    ZLDN (WSDDN + FRCNN)~\cite{zhang2018zigzag} \rule[0mm]{0mm}{3mm}& 47.6 & 42.9\vspace{0mm}\\
    ML-LocNet (WSDDN + FRCNN)~\cite{zhang2018ml} \rule[0mm]{0mm}{3mm}& 49.7 & 43.6\vspace{0mm}\\
    PGE (OICR + FRCNN)~\cite{zhang2018w2f} \rule[0mm]{0mm}{3mm}& 51.7&47.3\vspace{1mm}\\\hline
    Ours + FRCNN\rule[0mm]{0mm}{3mm}& 51.4 &{\bf 48.1}\vspace{0mm}\\
    PGE (Ours + FRCNN)\rule[0mm]{0mm}{3mm}&{\bf 52.1}& 47.9\vspace{1mm}\\\hline
  \end{tabular}
  }
    \label{map_multi}
    }
\end{table}

\begin{table}[t]
  \centering
  \caption{Effect of each labeling method to mAP (\%) on PASCAL VOC 2007.}
    {\tabcolsep=1.12mm
    {\scriptsize
  \begin{tabular}{l|DD}
    \multicolumn{3}{l}{} \vspace{0.1mm}\\\hline
        method\rule[0mm]{0mm}{3mm}&mAP&CorLoc\vspace{1mm}\\ \hline\hline
    baseline (OICR~\cite{tang2017multiple}) \rule[0mm]{0mm}{3mm}& 41.2 & 60.6\vspace{0mm}\\
    CAP labeling \rule[0mm]{0mm}{3mm}& 45.6 & 66.6\vspace{0mm}\\
    SRN labeling \rule[0mm]{0mm}{3mm}& 45.1 & 63.4\vspace{0mm}\\
    CAP and SRN labeling \rule[0mm]{0mm}{3mm}&{\bf 47.6}&{\bf 66.7}\vspace{1mm}\\\hline
  \end{tabular}
  }
    \label{map_abl}
    }
  \vspace{-2mm}
\end{table}

Using our localization result, we train a Fast R-CNN~\cite{girshick2015fast} (FRCNN) detector.
The result is shown in Table~\ref{map_multi}.
The first to third methods employ the predicted top-scoring region as the pseudo ground truth;
the fourth to the sixth methods focus on how to train a FRCNN detector effectively using the localization result.
Following the former methods, we train a FRCNN detector using the top-scoring region by our method, whose result is shown as Ours + FRCNN.
In addition, we apply Pseudo Ground-truth Excavation (PGE)~\cite{zhang2018w2f}, which is a previous state-of-the-art method for mining more accurate and tighter boxes instead of only one top-scoring box.
As shown in Table~\ref{map_multi}, our method obtains the highest score when combined with PGE for VOC 2007,
and outperforms the previous methods with and without PGE for VOC 2012.

\subsection{Ablation experiments}
\vspace{-2mm}
We conduct extensive ablation experiments to analyze our method. All ablation experiments are conducted on the VOC 2007 dataset.

\vspace{-2mm}
\noindent
\newline
{\bf Contribution of each labeling~~}
Our method is composed of CAP labeling and SRN labeling.
We investigate how much each method contributes to the improvement and show the results in Table~\ref{map_abl}.
Each method can improve the performance, and we can obtain even greater improvement by combining both labeling methods.

\vspace{-2mm}
\noindent
\newline
{\bf Context classification~~}
CAP labeling is based on the hypothesis that the context classification loss differs depending on how much of the object is covered by the region.
Here, to verify the hypothesis, we visualize the training loss curve.

\begin{figure}[t]
 \centering
 {\includegraphics[width=0.9\hsize]{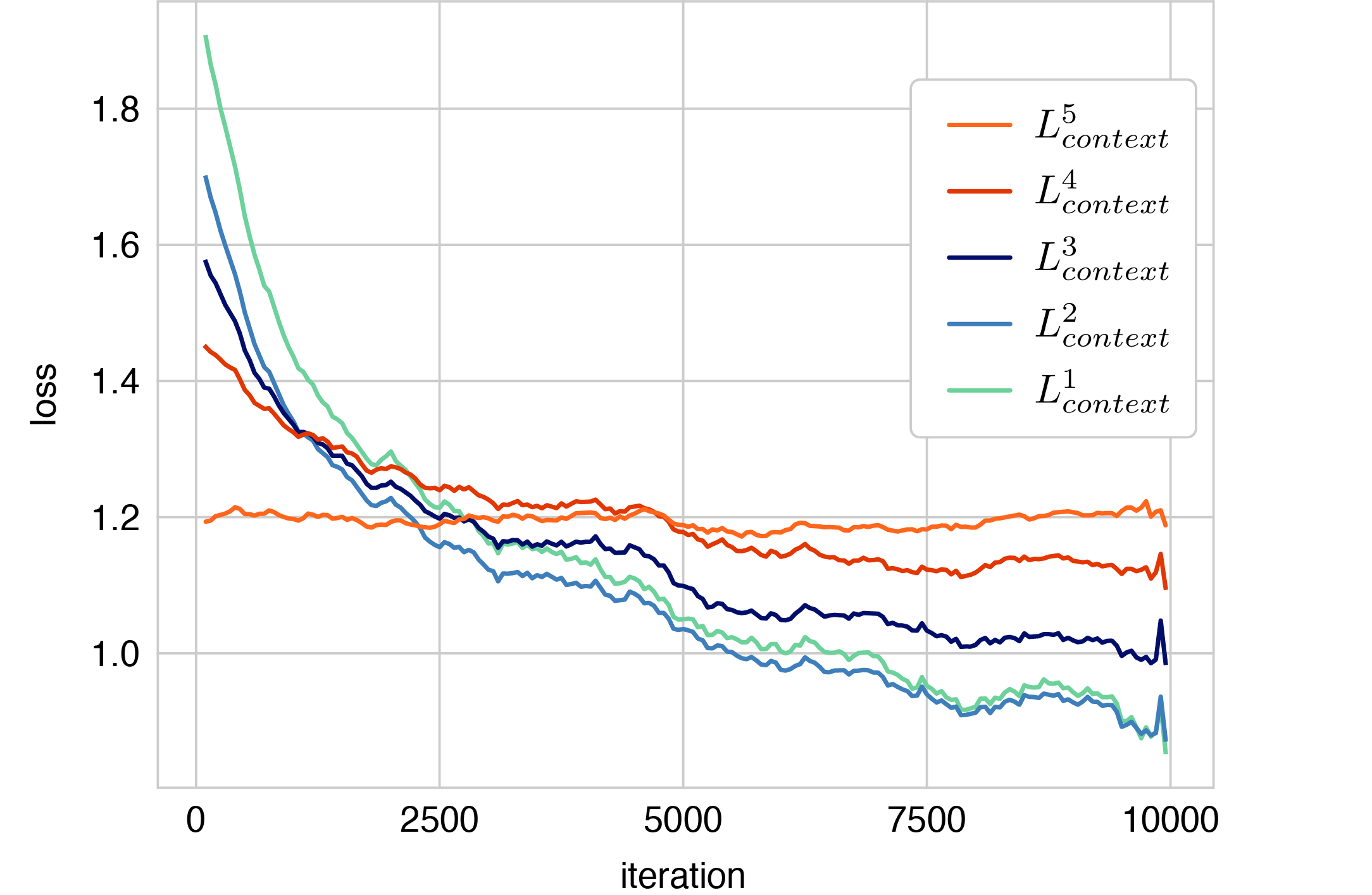}}
 \caption{Loss curve with different coverage of the object while training on the PASCAL VOC 2007 dataset.
 Through the training process, losses with low coverage decrease, while losses with high coverage do not decrease.\label{fig:loss_curve}
 \vspace{-2mm}
 }
\end{figure}

In order to divide regions according to the coverage of the object, we define the following sets,
\begin{equation}
  \mathcal{S}_i = \{(c, r_j)~|~0.2(i-1) \leq {\rm \scalebox{0.9}[1]{coverage}}(c, r_j) \leq 0.2i, y_c = 1\},
\end{equation}
where $i \in \{1, 2, 3, 4, 5\}$ and ${\rm coverage}(c, r_j)$ is a function calculating how much of the ground truth box of the class $c$ object is covered by the region $r_j$.
The context classifier is trained to minimize the loss Eq.~(\ref{eq:context_loss}).
To investigate the relationship between the coverage of the object and the training loss, we define the following loss,
\setlength{\abovedisplayskip}{5pt}
\setlength{\belowdisplayskip}{2pt}
\begin{equation}
  {L}_{context}^i = - \frac{1}{|\mathcal{S}_i|} \sum_{c, r_j \in \mathcal{S}_i} {\rm log}~p_{cj}.\label{eq:loss_visual}
\end{equation}
As we consider only the classes contained by the image, ${\rm log}(1-p_{cj})$ is not calculated.
Note that the objective function is loss Eq.~(\ref{eq:context_loss}) and  ${L}_{context}^i$ is only used for visualization.

The change of each ${L}_{context}^i$ during the training process is shown in Figure~\ref{fig:loss_curve}.
The training loss of the boxes covering only some parts of the object (${L}_{context}^1$, ${L}_{context}^2$) decrease through the training process.
On the other hand, the loss of the boxes covering most of the object (${L}_{context}^5$) does not decrease.
This result shows that the context classification loss can be used to find boxes covering the whole object.

\begin{spacing}{0.997}
\vspace{-2mm}
\noindent
\newline
{\bf Context classification and simple mask-out~~}
In some previous methods~\cite{bazzani2016self,li2016weakly}, a classifier is trained with image-level annotations,
and regions whose mask-out drops the classification confidence are defined as the object.
We refer to such methods as {\it simple mask-out}.
Here, we compare our context classification and the simple mask-out approach.

To perform the simple mask-out, we first train a standard classifier, which has the same architecture as the context classifier except for the mask-out.
Let an input image be $X$, the image label vector be ${\mathbf{Y}} = [y_1, ..., y_C]$, and the class probability be ${\mathbf{\overline{p}}} = [{\overline p}_1, ..., {\overline p}_C]$.
The classifier is trained to minimize the following classification loss,
\setlength{\abovedisplayskip}{2pt}
\setlength{\belowdisplayskip}{2pt}
\begin{equation}
  {L}_{simple} = - \sum_{c=1}^{C}\{y_{c}\log \overline{p}_{c} + (1-y_{c})\log (1-\overline{p}_{c})\}.\label{eq:classification_loss}
\end{equation}
After training, we mask-out the CNN feature corresponding to each region $r_j$ in the same way as context classification and obtain the probability $\overline{p}_{cj}$.
In Bazzani et al.~\cite{bazzani2016self} and Li et al.~\cite{li2016weakly}, mask-out is performed on the input image, but it requires forwarding for each region and is very time-consuming.
For the same CNN forwarding time, we mask-out not the input image but the CNN feature.

\begin{figure}[t]
 \centering
 {\includegraphics[width=\hsize]{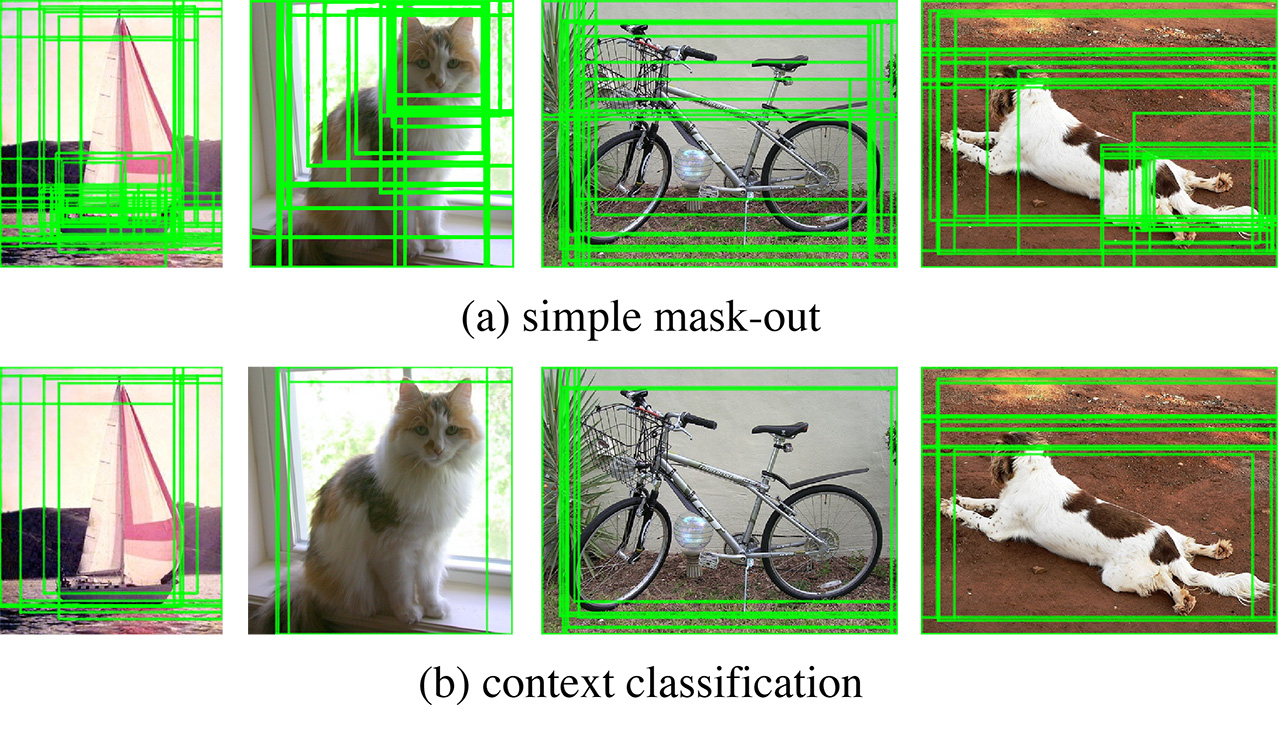}}\vspace{-2mm}
 \caption{Discovery result of regions covering the whole object by each method.\label{fig:remained_regions}
 }
\end{figure}

\begin{table}[t]
  \centering
  \caption{Comparison between our method and our method whose context classification is replaced with the simple mask-out when using the PASCAL VOC 2007 dataset.}
  \vspace{-2mm}
    {\tabcolsep=1.12mm
    {\scriptsize
  \begin{tabular}{l|DD}
    \multicolumn{3}{l}{} \vspace{0.1mm}\\ \hline
        method\rule[0mm]{0mm}{3mm}&mAP&CorLoc\vspace{1mm}\\ \hline\hline
    simple mask-out \rule[0mm]{0mm}{3mm}& 47.1 & 64.9\vspace{0mm}\\
    context classification \rule[0mm]{0mm}{3mm}& {\bf 47.6} & {\bf 66.7}\vspace{1mm}\\\hline
  \end{tabular}
  }
    \label{fig:w/simple_mask-out}
    }
    \vspace{-0mm}
\end{table}

To compare context classification and simple mask-out, we show regions whose $p_{cj}$ or $\overline{p}_{cj}$ are lower than the threshold $P_t$ in Figure~\ref{fig:remained_regions}.
Although with simple mask-out, the confidence of the classifier decreases when some parts of the object are covered,
the confidence of our context classifier drops only when the whole object is covered.
As a result, we can obtain regions covering the whole object.

We train our model by replacing context classification with simple mask-out (Table~\ref{fig:w/simple_mask-out}).
In both metrics, CorLoc and mAP, the method using context classification achieves better performance.
This result demonstrates the effectiveness of our context classification.

\vspace{-2mm}
\section{Conclusions}
\vspace{-2mm}
In this study, we address weakly supervised object detection.
As a typical method to train a detector with image-level annotations, the detector and the instance-level labels are updated iteratively.
In order to achieve more efficient iterative updating, we focus on the instance labeling problem, a problem of which label should be annotated to each bounding box based on the last localization result.
We improve instance labeling in two ways.
First, we label boxes covering the whole object as positive, being aware that the context classification loss differs according to the coverage of the object.
Second, we introduce a spatial restriction to avoid labeling other objects as negative.
Experiments show that our method achieves significant improvement.

\vspace{2mm}
\noindent
{\bf Acknowledgement~~} A part of this research was supported by JST-CREST~(JPMJCR1686) and the Grants-in-Aid for Scientific Research (19K22863).
\end{spacing}

{\small
\bibliographystyle{ieee_fullname}
\bibliography{egbib}
}
\end{document}